\documentclass[letter, 10 pt, conference]{ieeeconf}
\IEEEoverridecommandlockouts                              % This command is only needed if
\usepackage{amsmath}
\usepackage{amssymb}
\usepackage{mathtools}
\usepackage{graphicx}
\usepackage{float}
\usepackage{array}
\usepackage{tikz}
\usepackage{listings}
\usepackage{soul}
\usepackage{multirow}

\PassOptionsToPackage{hyphens}{url}
\usepackage{hyperref}
\usepackage{graphicx}
\usepackage{cite}
\usepackage{dsfont}
\usepackage{mathtools,etoolbox}
\usepackage[ruled, linesnumbered]{algorithm2e}
\usepackage[none]{hyphenat}
\usepackage[utf8]{inputenc}
\usepackage[english]{babel}

\usepackage{amsthm}
\usepackage{xfrac}
\usepackage{nicefrac}
\usepackage{booktabs}
\usepackage[switch, pagewise]{lineno}
\usepackage{nicefrac}
\usepackage{flushend}
\hypersetup{colorlinks=true,urlcolor=red,citecolor=red}

\usepackage[placement=top]{background}

\backgroundsetup{
        placement=top,
        position={8cm,2cm}}
\usepackage[final]{pdfpages}

\SetBgScale{1}
\SetBgContents{\parbox{16cm}{\Large \centering This paper is under review for publication at the IEEE International Conference on Robotics and Automation (ICRA),  London, 2023. \textcopyright IEEE}}
\SetBgColor{gray}
\SetBgAngle{0}
\SetBgOpacity{0.9}

\title{\LARGE \bf WorldGen: A Large Scale Generative Simulator}
\author{Chahat Deep Singh$^1$, Riya Kumari$^1$, Cornelia Ferm{\"u}ller$^1$, Nitin J. Sanket$^2$, Yiannis Aloimonos$^1$ % <-this % stops a space
\thanks{This work was supported in parts by Northrop Grumman (N00014-17-1-2622, SMA 1540917), Army Research Laboratory under a cooperative agreement with the University of Maryland (ArtIAMAS), Army Cooperative Agreement (W911NF2120076) and NSF Grant (OISE 2020624). Chahat Deep Singh was funded by Ann G. Wylie fellowship at the University of Maryland, College Park.}
\thanks{
$^{1}$Perception and Robotics Group, University of Maryland Institute for Advanced Computer Studies, University of Maryland, College Park, MD 20742, USA. Emails: \texttt{\{chahat, rkumari, fer, jyaloimo\}@umd.edu.}}
\thanks{
$^{2}$Robotics Engineering, Worcester Polytechnic Institute, MA 01609, USA. Email: \texttt{nsanket@wpi.edu}}}
\begin{document}
\makeatletter
% \g@addto@macro\@maketitle{
% \begin{figure}[H]
%   \setlength{\linewidth}{\textwidth}
%   \setlength{\hsize}{\textwidth}
%     \centering
%     \includegraphics[width=0.9\textwidth]{Images/Banner3.png}
%     \caption{Generative ability of WorldGen: (a) Comparison between Google Street View (left) and the same street in WorldGen (right), (b) Comparison of Google Maps satellite image vs. WorldGen top view, (c) Collection of 3D objects in motion, (d) Object fragmentation,(e) Annotation from left to right: depth, optical flow, surface normals, stereo anaglyph, image segmentation, event frame.}
%     \label{fig:Banner}
%     \end{figure}
% }
\maketitle

%%%%%%%%%%%%%%%%%%%%%%%%%%%%%%%%%%%%%%%%%%%%%%%%%%%%%%%%%%%%%%%%%%%%%%%%%%%%%%%%%%%%%
\begin{abstract}
In the era of deep learning, data is the critical determining factor in the performance of neural network models. Generating large datasets suffers from various difficulties such as scalability, cost efficiency and photorealism. To avoid expensive and strenuous dataset collection and annotations, researchers have inclined towards computer-generated datasets. Although, a lack of photorealism and a limited amount of computer-aided data, has bounded the accuracy of network predictions.

To this end, we present WorldGen -- an open source framework to autonomously generate countless structured and unstructured 3D photorealistic scenes such as city view, object collection, and object fragmentation along with its rich ground truth annotation data. WorldGen being a generative model gives the user full access and control to features such as texture, object structure, motion, camera and lens properties for better generalizability by diminishing the data bias in the network. We demonstrate the effectiveness of WorldGen by presenting an evaluation on deep optical flow. We hope such a tool can open doors for future research in a myriad of domains related to robotics and computer vision by reducing manual labor and the cost of acquiring rich and high-quality data. The project page is available at \hyperlink{http://prg.cs.umd.edu/WorldGen}{http://prg.cs.umd.edu/WorldGen}

% The abstract goes here...
% Generative Scene
% Simulator for rendering adverse condition (Like Thermal and HDR)
\end{abstract}

\section{Introduction}
High-quality image data is pivotal for accurate deep learning models. For certain tasks, it is even more important than neural architectures and hyperparameters. Data collection, cleaning, and annotation have become a nightmare lately generally costing millions of dollars. This is majorly due to challenges in data diversity, image quality, pixel-perfect manual annotations, as well as licensing \cite{asano2021pass}, and security concerns. This is exacerbated by the fact that writing instructions for manual annotation are not a trivial task as it is not always possible to explain concisely the requirements. To this end, we propose a universal framework to generate countless 3D scenes in different environments along with annotated images for various autonomous tasks for cars, drones, and indoor robots. This would enable deep learning frameworks to gather photorealistic data much more efficiently in terms of cost, speed, labor and variety.

On the other side of the spectrum, synthetic data and simulators today are commonly used but are very limited as they suffer from \textit{scalability} and/or \textit{photorealism}. Researchers and practitioners have extensively used these data as a benchmark to evaluate the depth, optical flow, etc. due to their \textit{perfect} ground truth annotation. Most of these simulators do not support a realistic camera model or ray-tracing ability to render high-quality photorealistic images. Although, few approaches have proven to narrow down the generalization gap using synthetic data \cite{dosovitskiy2015flownet, mayer2016large} for geometric-based quantities such as depth and optical flow as opposed to texture-centric approaches like semantic segmentation.

\begin{figure}[t!]
    \centering
    \includegraphics[width=1.0\columnwidth]{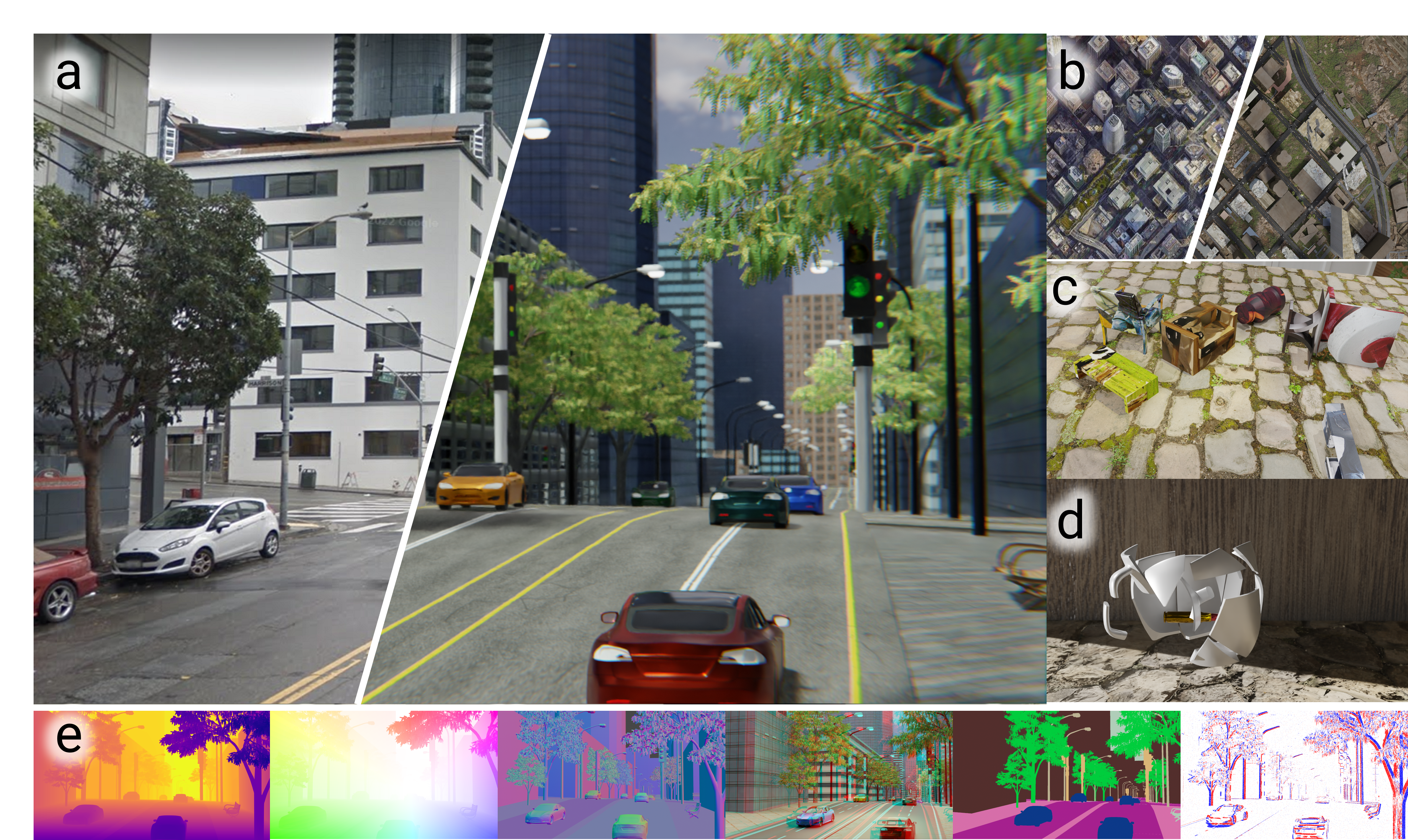}
    \caption{Generative ability of WorldGen: (a) Comparison between Google Street View (left) and the same street in WorldGen (right), (b) Comparison of Google Maps satellite image vs. WorldGen top view, (c) Collection of 3D objects in motion, (d) Object fragmentation,(e) Annotation from left to right: depth, optical flow, surface normals, stereo anaglyph, image segmentation, event frame. \textit{All the images in this paper are best viewed in color on a computer screen at 200\% zoom}.}
    \label{fig:Banner}
    \end{figure}
    
    \begin{table}[t!]
    \centering
    \caption{Comparison of the different simulation environments.}
    \noindent\begin{tabular}{{l|}*{5}{c}}
         % \hline
        \toprule
         Name & Rendering & GI & Physics & Scaling\\
         \midrule
         UnrealCV \cite{qiu2016unrealcv} & UE4 & $\times$ & UE4 \ & $\times$\\
         iGibson \cite{igibson} & PyRender & $\times$ & PyBullet & \checkmark \\
         Omnidata \cite{omnidata} & Blender & \checkmark & -- & $\times$ \\
         Blenderproc \cite{blenderproc} & Blender & \checkmark & Bullet & $\times$ \\
         AirSim \cite{airsim} & UE4 & $\times$ &  AirSim & $\times$  \\
         CARLA \cite{carla} & UE4 & $\times$ &  UE4 & $\times$  \\
         Kubric \cite{kubric} & Blender & \checkmark & PyBullet & \checkmark\\
         WorldGen \textit{(Ours)} & Blender & \checkmark & Bullet & \checkmark\\
         \bottomrule

    \end{tabular}
    \\\tiny{$^*$ Only Blender rendering engine supports ray-tracing with OptiX denoiser for photorealistic images; others use rasterization. GI: Full Global Illumination Support; Physics Engine Used; Scaling: Scalable to generate large datasets.}
    \label{tab: comparison}
\end{table}

To reduce this sim-to-real gap even further, a large amount of synthetic scene generation with wide variability is required. Creating a large number of synthetic scenes suffers from either manually creating high-quality 3D assets or bringing together these assets in a structured way to form a scene. Thus, a scalable simulation environment becomes a necessity for neural networks to improve the quality of the prediction models. \cite{kubric} and \cite{igibson} \ref{tab: comparison} demonstrated the power of scalable data by providing an in-depth analysis of the prediction using their data for training. In this work, we present a scalable framework to automatically generate structured and unstructured environments for perception and robotics applications without any need for manual effort. Fig. \ref{fig:Banner} demonstrates the ability of WorldGen to autonomously generate structured cities, object datasets and object fragments without any requirement for manual efforts.

\subsection{Related Work}

In the last five years, there has been a boom in self-driving car simulators for perception, planning and control \cite{summit, airsim, carla}. One of the major drawbacks of such simulators is the limitation in scalability for data generation. Such simulators are shipped with a limited number of scenes (or towns). This is because it requires intensive manual labor to generate these 3D environments for better scalability and generalizability. Scalability in  these simulators is present for traffic simulations but not for perception tasks. For instance, CARLA \cite{carla} library includes only 40 different building models. Our method relies on 3D structures of buildings from satellite semantics and 3D maps to build countless numbers of building models.

Learning geometric annotations such as an optical flow that relies substantially on motion parameters (rather than textures) have been studied extensively in the past decade. Unlike monocular depth estimation and instance segmentation, the prediction of such geometric quantities through convolutional neural networks generalizes well across different domains. For such predictions, synthetic datasets have widely been used \cite{mayer2016large, dosovitskiy2015flownet, scharstein2002taxonomy, sintel} with a limited number of objects and textures. Recent advancements in scalable scene generation environments like Kubric \cite{kubric} have shown improvements in optical flow prediction due to more variability in data generation. 

Another common problem in vision and robotics is segmenting and tracking independently moving objects. It is a key process to understand the dynamics of the scene structure for navigation tasks. Applications such as dodging any malicious objects on drones \cite{evdodge, falanga2020dynamic} and tracking objects using RGB and event cameras in motion \cite{parameshwara20210, wedel2009detection} have been well examined. Also, from the virtue of archaeology, it is an important task to reconstruct a broken pot from unordered and mix fragment collection \cite{hong2021structure}.
In this paper, we provide a framework to generate countless 3D scenes with texture, structure and lighting variability for the aforementioned applications. We summarize our key contributions next.

\subsection{Key Contributions}
\begin{itemize}
\item We introduce WorldGen, a generative simulator for creating different environments commonly encountered in real-world robotics and computer vision tasks
\item WorldGen is scalable and can generate an infinite amount of photorealistic data with variations to not only object placements but also object shape, texture, lighting, camera and motion properties.
\item WorldGen supports realistic camera distortions such as barrel distortion, chromatic aberration, and camera aperture among other computational photographic properties.
\item We demonstrate WorldGen's utility in learning more accurate optical flow due to the virtue of better data generation ability.
        % \item In this paper, we propose a generative simulator to create different environments for various perception, planning, and learning tasks using only satellite data and other assets like objects, materials, etc.
\end{itemize}

The paper is organized as follows: Sec. \ref{sec:Worldgen} presents how our WorldGen generative simulator is constructed, its design principles and details. Next, Sec. \ref{sec:Applications} presents how WorldGen can be useful for a myriad of robotics and computer vision applications. Finally, we conclude the paper in Sec. \ref{sec:conc} with parting thoughts for future work.

\begin{figure*}[h!]
    \centering
    \includegraphics[width=0.8\textwidth]{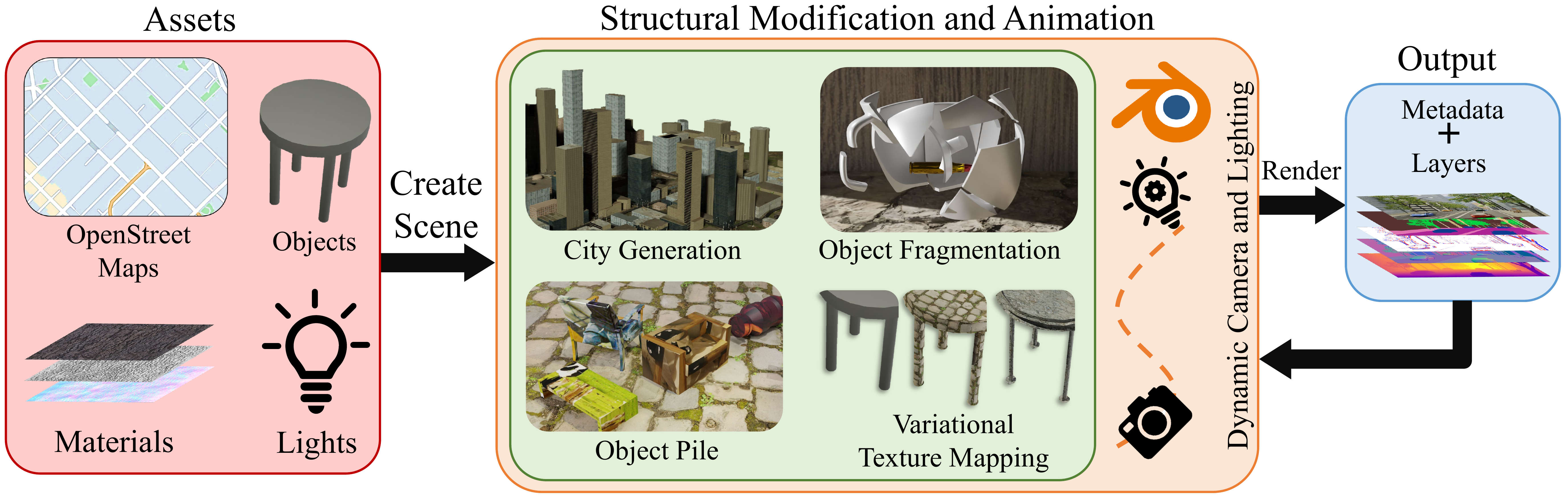}
    \caption{An overview of WorldGen Framework: (a) Assets: Loads the assets such as maps, objects, materials etc. into WorldGen environment, (b) Structural Modification and Animation: Modifying the texture maps and applying physics and motion models on different objects in the scene, (c) Rendering: Generates rich ground truth data with the desired metadata (time, frame number, camera intrinsic and extrinsic properties).}
    \label{fig:Overview}
\end{figure*}

\section{WorldGen Generative Simulator}
\label{sec:Worldgen}
WorldGen is a high-level open-source python library to generate an unlimited amount of synthetic data. It serves as an ease-of-use platform to generate visual data for simulating self-driving cars, autonomous drones, object segmentation, active vision, motion segmentation, tracking, computational photography and more. Our key contribution is an API to build generative environments and streamline the process of generating synthetic data by lowering the usage difficulty barrier for researchers and practitioners alike. WorldGen is built around Blender$^\text{TM}$, a free and open-source 3D creation suite, to generate synthetic data like city maps, a collection of moving objects, and object fragmentation. Currently, our framework utilizes \texttt{Bullet} physics engine mainly for object collisions, gravity, friction, and other force fields. Fig. \ref{fig:Overview} shows an overview of WorldGen. We build WorldGen around the central concept of scalability and speed. We discuss different parts  and a number of details 
that are used to build WorldGen next.

\subsection{Environment}
\label{subsec:environment}
Our framework is structured in three different stages: (a) Loader: loading assets like objects, materials and textures, lights and/or map information to generate a structured 3D environment, (b) Structural Modification and Animation: this generates structurally different objects by varying texture maps (explained in Sec. \ref{subsection:UVMaps}), then applies physics and motion models to objects, camera and lights and (c) Rendering: outputs rendered frames and its annotation along with desired metadata (like time, frame number, camera intrinsic, and extrinsic).

\textit{Rendering:} Our framework utilizes Blender's \textit{Cycles} -- a ray-trace based production render engine. To avoid slow renders, the number of samples (or path-traced for each pixel) is usually set to a lower value that produces \textit{graininess} in the output renders. We utilize NVIDIA OptiX\textsuperscript{TM} \cite{optix}, a recurrent denoising autoencoder to reduce the \textit{graininess} (noise) without a need for more rendering iterations. Furthermore, our framework also utilizes both full Global Illumination (GI) for photo-realistic renders and fast GI approximation for faster renders.
% Global Illumination (GI) support: https://drive.google.com/drive/folders/1nIL-bGTHaKDeS2rBOxwDNACq6BhpASZh

\subsection{Texture Mapping}
\label{subsection:UVMaps}
\begin{figure}[h]
    \centering
    \includegraphics[width=0.9\columnwidth]{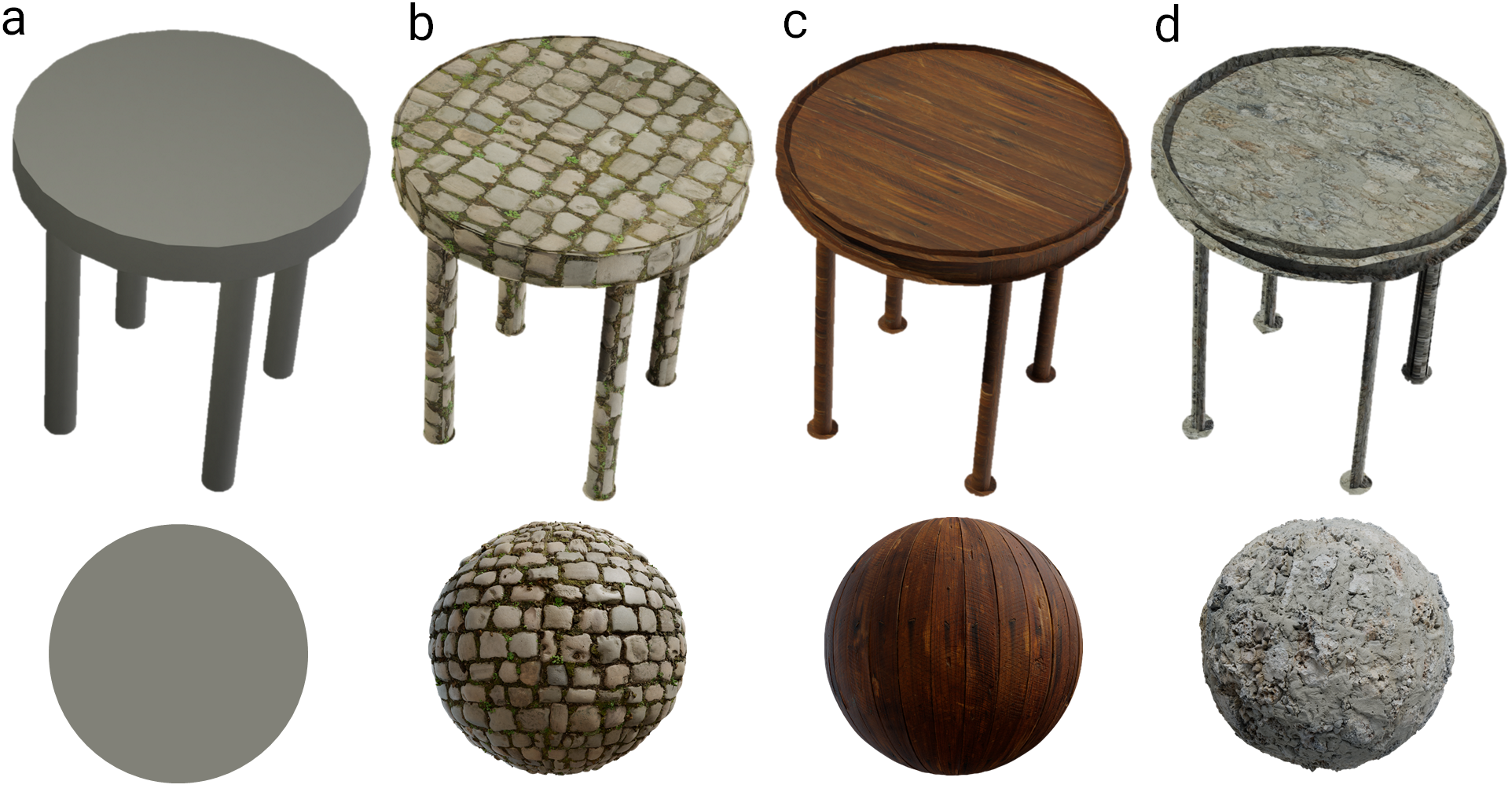}
    \caption{Mapping textures to a round table. Top row: Rendered Output, Bottom row: Sample textures projected on a sphere. (a) Barebone 3D model, (b)-(d) Different Textures applied on (a). \textit{Note: Variational mapping models change the structure of the 3D objects in different renders (notice the legs on the chair). Here, the Gaussian noise in $(d)>(c)>(b)$.}}
    \label{fig:TextureMaps}
\end{figure}

WorldGen utilizes UV mapping \cite{levy2002least} which is a well-known process in computer graphics that projects a 2D image onto a 3D model's surface. Along with RGB images, open-source textures are often composed of displacement and normal maps in order to \textit{fake} the lighting of \textit{bumps} in the 3D model. These mapping techniques are used to re-detail a simplified or low-poly mesh without increasing the number of vertices, thus reducing the rendering time. By varying the strength of the displacement and normal maps, a structurally different variant of the same object can be rendered without a need to modify the 3D object (See Fig. \ref{fig:TextureMaps}). Thus, we modify these maps with additive white Gaussian noise before applying them to the 3D objects. Such variational texture mapping can contribute to the generation of thousands of \textit{similar} object renders from a small set of object meshes. This can be used to generate large amounts of datasets to predict geometric quantities (or annotations) like optical flow for better generalizability across different domains. For more details, read Sec. \ref{label:OpticalFlow}.

\subsection{Generative Models}
At the time of writing, WorldGen supports three different generative models: \textit{(1)} City Maps \textit{(2)} Object Pile, and \textit{(3)} Object Fragmentation.\\[-5pt]

\subsubsection{City Models}

\begin{figure}[t]
    \centering
    \includegraphics[width=0.9\columnwidth]{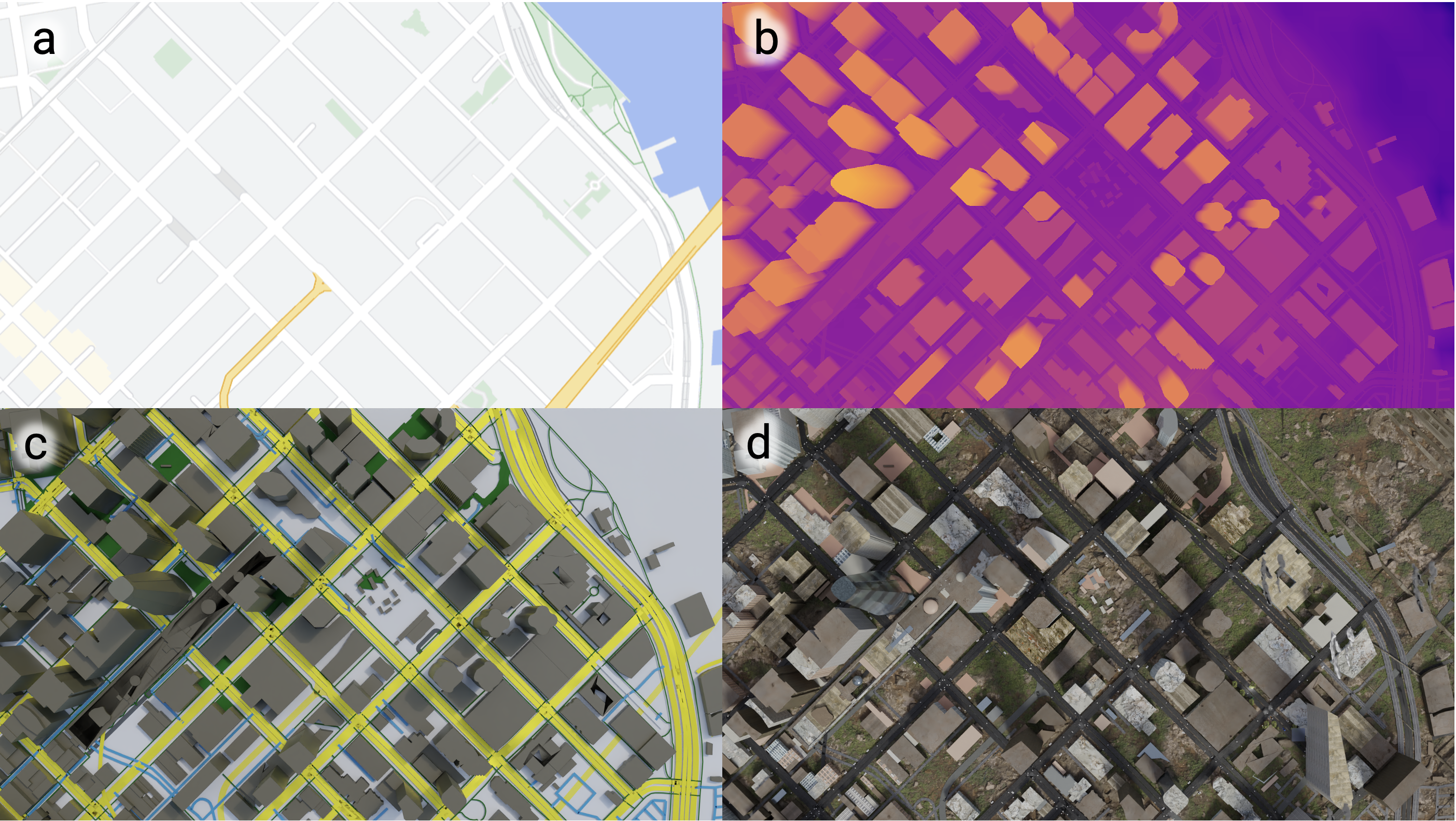}
    \caption{(a) OpenStreetView, (b) Depth Map, (c) 3D Model View Generated by WorldGen and (d) Final Rendered View}
    \label{fig:Nerf}
\end{figure}

\begin{figure}[t]
    \centering
    \includegraphics[width=0.9\columnwidth]{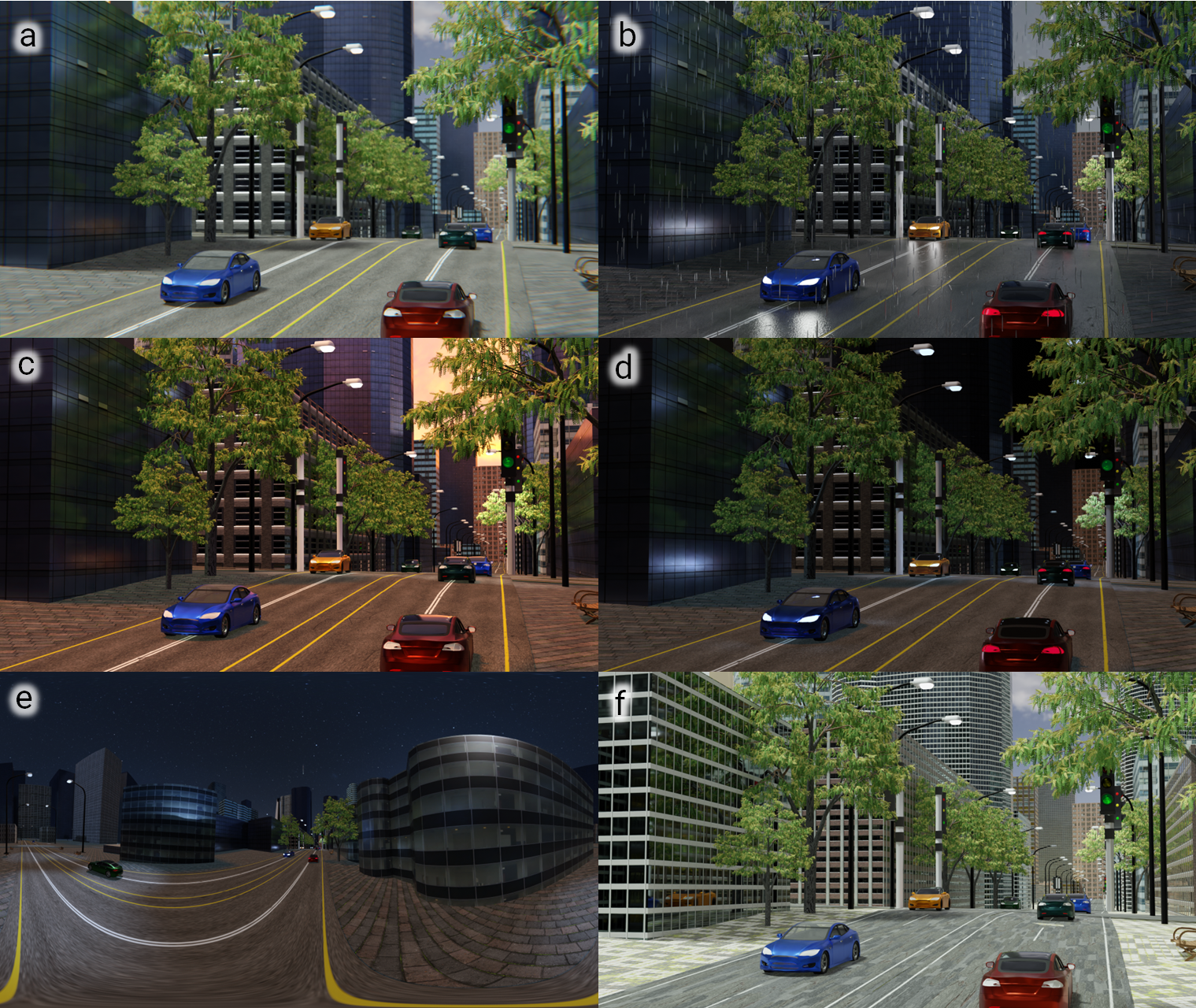}
    \caption{City environment in different weather and time of the day: (a) Day, (b) Night with rain, (c) Dawn and (d) Night without rain, (e) panoramic view of the city and (f) demonstrates the generative ability of WorldGen by changing the textures of the entire scene while keeping the same structure.}
    \label{fig:DifferentEnvironments}
\end{figure}

Our framework utilizes OpenStreetMaps (OSM), a crowd-sourced project that contains semantic labels and 3D terrain maps from the satellite perspective. Using latitude and longitude as the input, semantic and 3D terrain information is imported into our framework using \cite{blenderosm}. Using semantic information like buildings, water bodies, forests, vegetation, roads, highways, pedestrian pathways and railways, a set of relevant assets are imported from open source libraries (see Sec. \ref{subsec:Assets}) along with its appropriate material. Furthermore, we can deploy different roof structures such as flat or gable on various buildings. the semantic data from OSM also contains building roof information, allowing our framework to deploy a \textit{realistic} roof structures (flat or gable) on the respective buildings. Assets like radio antennas, air-conditioning vents, and chimneys are randomly distributed over the roof surface. For other generative solid objects like roads, 3D meshes are created in the environment using the vertices from the semantic data. Furthermore, lights are imported to the scene with GI support for different types of weather and daytime scenarios to get a photo-realistic visualization (Fig. \ref{fig:DifferentEnvironments}). Using basic morphological and vector operations, the road intersections are estimated to deploy traffic lights and stop signs near these road crossings. Other commonly available street objects like trees, benches, and street lights are structurally distributed near the pedestrian pathways. Fig. \ref{fig:Banner} shows the final rendered output of our city environment, complete with realistic distortions present in camera images such as chromatic aberration. \\[-5pt]

\subsubsection{Object Pile}
WorldGen's ability to generate a collection of moving objects in space is closest in scope to Kubric \cite{kubric} and BlenderProc \cite{denninger2019blenderproc}. The key advantage of our framework over Kubric is the flexibility to generate large-scale datasets due to variational texture mapping that can generate different structural variations of the same 3D models. Fig. \ref{fig:Banner}{\color{red}b} shows the output of an object pile environment. WorldGen supports a variety of control in physics such as collision shapes using convex hulls, 3D meshes, boxes, cylinders, etc. with varying collision margins, friction, and coefficient of restitution for allowing more control in the data generation process. Furthermore, apart from the standard gravitational force, our framework deploys other force fields such as wind and drag to influence the object's motion which reflects reality much more and would be useful in learning perception modules for realistic scenarios.\\[-5pt]

% Real Snow: https://docs.blender.org/manual/en/latest/addons/object/real_snow.html
% https://blenderartists.org/t/problems-with-cell-fracture-and-the-game-engine/581626/3
\subsubsection{Object Fragmentation}
Another generative model of our framework is to break or \textit{fragment} any 3D mesh into a user-defined number of Voronoi cell fractures. We utilize \cite{cell_fracture} to generate the specified number of 3D meshes from the parent mesh. The object fragments into child particles upon contact with another rigid body (active or passive) or any other non-uniform force field. The rendered output is a video sequence from the camera sensors with semantic information of each individual fragment. Fig. \ref{fig:Banner}{\color{blue}c} shows the simulation environment of the scene where a bullet is colliding with a cup, thereby shattering it into a number of pieces. Such an approach is potentially useful for assessing damage due to collisions. \\[-5pt]

\subsection{Sensors}
WorldGen renders RGB images along with the annotation data such as depth map, optical flow, stereo images, semantic map, and surface normal map (See Fig. \ref{fig:Banner}) which are rendered directly through the Blender Cycles engine. For event camera renders, we utilize a simple event camera model \cite{sanket2021evpropnet, evdodgenet} to generate events at a location $\mathbf{x}$ when
\begin{equation}
  \Vert \log \left(\mathcal{I}_t\left(\mathbf{x}\right) \right)- \log \left(\mathcal{I}_{t+1}\left(\mathbf{x}\right) \right)\Vert_1 \ge \tau
\end{equation}

where $\tau$ is a user-defined threshold, $\mathbf{x}$ is the pixel location and $\mathcal{I}_{t}$ represents a grayscale image captured at time $t$. An \textit{event-frame} is generated using the above model along with additive Gaussian noise. Alternatively, a more realistic event model \cite{v2e, rpg_vid2e} can be adapted to our framework. For smaller integration time in generating event frames, we remap the animation timeline to $\mathbb{N}$ times more frames to give a slow-motion effect to simulate a visual high-speed camera from which events are generated.

Multiple such sensors can be spawned in the environment to render outputs from different camera pose at the same time. WorldGen takes 6 degree-of-freedom camera extrinsic parameters to render outputs from different cameras. This setup can be used to simulate a sensor suite like the ones typically found on a self-driving car or an autonomous drone.\\[-5pt]

\textit{Camera Properties:} WorldGen supports both dynamic camera trajectories as well as variation in camera intrinsic properties -- dynamic focal length, depth of field, camera aperture radius, or variable camera baseline in the case of stereo output to simulate and generate data for previous works like \cite{morpheyes, pueyo2022cinempc}. These properties can be temporally modified by using simple linear interpolation between defined key-frames or Bézier curves. WorldGen uses the same method to dynamically move the solid objects in the scene. Also, due to variability in camera focal length, our framework supports both fisheye and equirectangular projection to render full 360-panorama images (Fig. \ref{fig:DifferentEnvironments}{\color{red}{e}}). Furthermore, WorldGen also supports real-world camera distortions for a more photorealistic render such as lens glare, chromatic aberrations, and barrel distortion. This is particularly useful for handling corner case scenarios which might aid in better sim-to-real generalization.

% \subsection{FOV, Depth of Field, Computational Photography Use and Other Lens Properties, Anamorphic}

\subsection{Lighting and Climate Conditions}
Currently, WorldGen supports three lighting conditions -- midday, sunset and night as well as 4 different weather conditions -- rain, cloudy, clear, and fog (Figs. \ref{fig:DifferentEnvironments}{\color{red}a} to \ref{fig:DifferentEnvironments}{\color{red}d}). Since WorldGen is Global Illumination (GI) enabled, the sunlight diffusion and reflection are photorealistic. Along with GI, it supports fast GI approximation for faster render time with a small loss in photorealism. The diffused sky radiation; the angle, power, and color temperature of the sun, and light sources such as street lights can be temporally varied.

\subsection{Assets}
\label{subsec:Assets}
The elegance of the WorldGen generative simulator is to use and modify existing 3D meshes, and materials to create new 3D environments. For the object pile and fragmentation scene generation, WorldGen imports objects, and materials by reading a text file that contains a list of wavefront \texttt{.obj} file path and material \texttt{.mtl} file path which makes it easier for the user to import any 3D object mesh into the WorlGen environment. We utilize a myriad of 3D object databases and image textures for incorporating a wide variety of possibilities. Furthermore, users can also import any custom objects/textures along with the ones described below: \\[-5pt]

\textbf{ShapeNetCore.v2 \cite{chang2015shapenet}: } ShapeNetCore.v2 contains about 51,300 unique 3D models across 55 object categories that are manually verified for classification and alignment annotations. We utilize Kubric's cleaned version of ShapeNet objects that fixes issues regarding auto-smoothing and backface culling that are present in the original ShapeNetCore objects.\\[-5pt]

\textbf{Google Scanned Objects (GSO) \cite{downs2022google}: } (CC-BY 4.0 License) GSO contains 1030 scanned objects and their associated metadata totaling ${\sim}13\text{GB}$. Since these models are scanned and not hand crafted in a 3D modeling tool, they realistically reflect real object properties.\\[-5pt]

\textbf{MS-COCO \cite{coco}: } Microsoft's COCO is a dataset that contains images of complex everyday scenes containing common objects in their natural context containing over 328k images across 91 object classes. We use these relatively lower-resolution images to warp around objects to modify their natural appearance.\\[-5pt]

\textbf{AmbientCG \cite{ambientcg}} is a large public domain resource with CC0 license for Physically Based Rendering (PRB) with 3D objects, materials and High Dynamic Range Images (HDRI) containing photo-scanned materials, displacement maps to ${\sim}1$ pixel accuracy with about 1760 assets with texture resolution up to $8K$.\\[-5pt]

\textbf{Polyhaven \cite{Polyhaven}} is a public library with a CC0 license with $500+$ pre-processed HDRI images which we resourced for backgrounds and lighting.\\[-5pt]

\textbf{cgbookcase \cite{cgbookcase}} is another CC0 license with $540+$ high-resolution PBR textures for common city structures such as different walls, roads, buildings, concrete, etc.\\[-5pt]

Apart from the aforementioned assets, we also utilized open-source libraries and Blender add-ons for pre-processing and rendering operations: \texttt{Blender-OSM}, \texttt{MapBox} and \texttt{object{\_}fracture{\_}cell} and \texttt{bpycv}.

% Blender-OSM: OpenStreeMap (OSM), MapBox, ArcGIS, bpycv %https://github.com/DIYer22/bpycv

% \begin{itemize}
%     \item Blender-OSM: OpenStreeMap (OSM)
%     \item MapBox
%     \item ArcGIS
%     \item bpycv %https://github.com/DIYer22/bpycv
% \end{itemize}

% \begin{itemize}
%     \item KuBasic
%     \item ShapeNetCore.v2 (cleaned)
%     \item GSO
%     \item Polyhaven
%     \item HDR\_VDS\_Dataset (Deep Chain HDRI)
%     \item MSCOCO
% \end{itemize}

\subsection{Design Principles}
 
\subsubsection{Modularity}
WorldGen is a highly modular framework as it is divided into three different elements as discussed in Sec. \ref{subsec:environment}. Each of the modules can be treated differently and can be replaced by a different third-party module. For instance, we can extend this work to generate a variety of human motions on a custom background using assets from 3D character rigging resources such as Adobe's Mixamo \cite{blackman2014rigging} to generate ground truth data annotations for human pose estimation. We plan to add character-rigged motion scene generation and data annotation with varying character motion to WorldGen.\\[-5pt]

\subsubsection{Ease of use}
The design philosophy of WorldGen is to make it easier for researchers, educators and practitioners to generate new scenes without needing any core knowledge of computer graphics concepts, modeling, animating, and rendering tools. WorldGen bridges this gap by presenting a simple high-level object-oriented Python API with Blender (and Bullet) working in the background.\\[-5pt]

\subsubsection{Open Source}
Scene and data generation codes will be released upon acceptance and will be free to use in academia. All the third-party assets used in WorldGen with Apache2.0, MIT and CC0 licenses enable researchers to generate different environments, render annotations and share all the data with the community to ensure reproducibility.\\[-5pt]

\subsubsection{Scalability}
WorldGen brings generative methods to build new structured environments without any need for manual labor. It can be used to generate environments anywhere between a small town and a large city with a compromise on GPU and access memory.

%
% https://console.cloud.google.com/storage/browser/kubric-public;tab=objects?pli=1&prefix=&forceOnObjectsSortingFiltering=false&pageState=(%22StorageObjectListTable%22:(%22f%22:%22%255B%255D%22))

\section{Applications}
\label{sec:Applications}
WorldGen is designed to generate different 3D environments and annotated data for robotics and computer vision applications. It is meant to be a one-stop framework for zero-shot generalization due to its variability in shapes, texture, and dynamic lighting; also minimizing the gap for sim-to-real transfer. It would also serve as a simulator for developing and testing planning and control algorithms due to its flexibility since it can use the current state and annotation as input for the next time stamp. To demonstrate the potential and flexibility of WorldGen, we discuss a set of common and challenging problem statements in robotics and computer vision.

% - Zero Shot Generalization
% - Sim2Real: Semantic to Real Translation (GAN, Transformers)

\subsection{Improvements in Optical Flow}
\label{label:OpticalFlow}
\begin{figure}[t!]
    \centering
    \includegraphics[width=0.9\columnwidth]{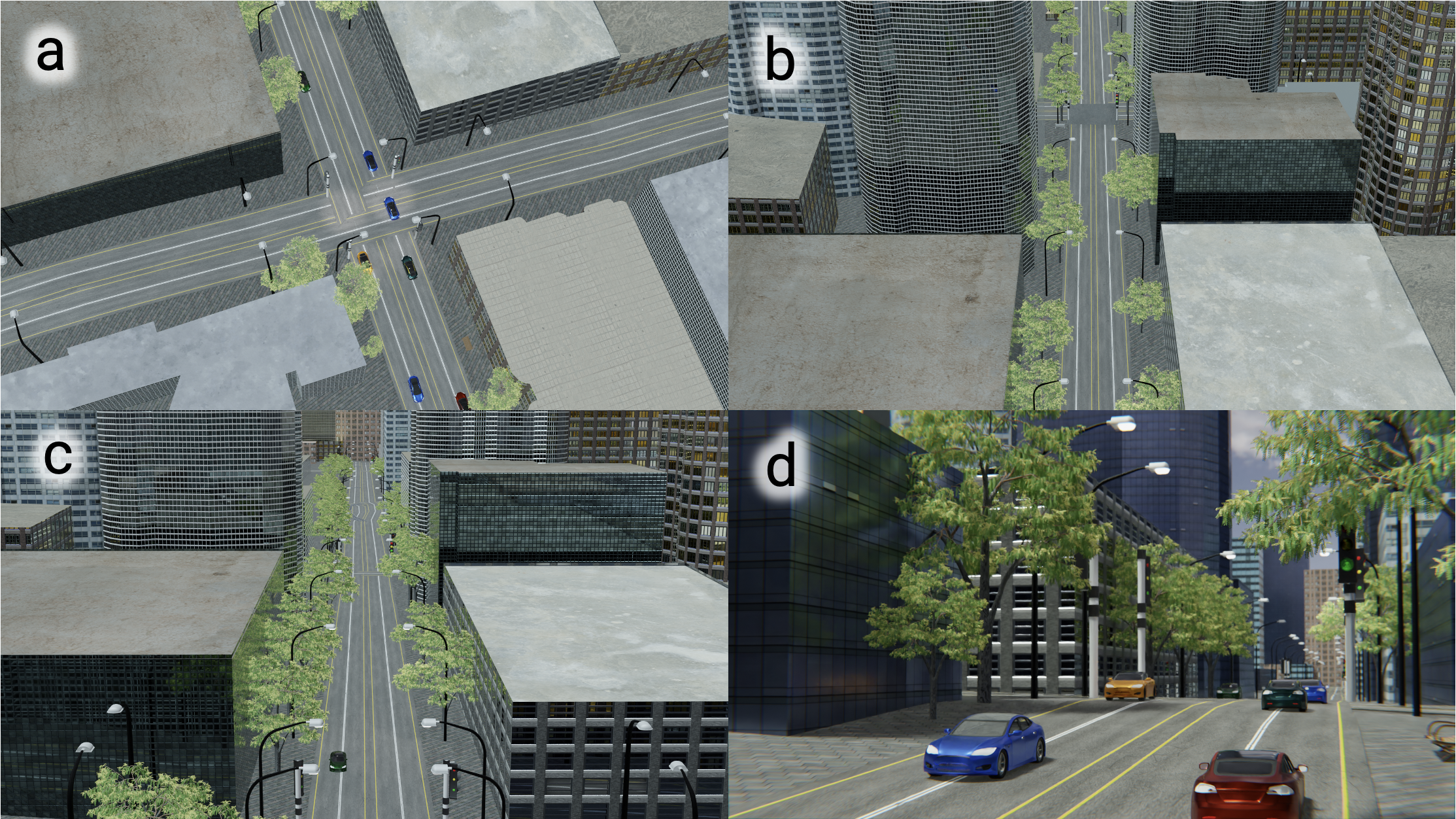}
    \caption{High resolution views generated by WorldGen from views at different altitudes with dynamic lighting, camera intrinsic, and extrinsic.}
    \label{fig:Nerf}
\end{figure}

Optical flow is one of the most fundamental quantities in computer vision and robotics which measures the 2D motion of each pixel between consecutive frames. Contrary to high-level computer vision applications like instance segmentation, ground truth cannot be reliably obtained from real-world data with human annotation. Recent advances in optical flow methods -- PWC-Net \cite{sun2018pwc}, RAFT \cite{teed2020raft} and GMFlow \cite{xu2022gmflow} all rely on synthetic data like MPI Sintel \cite{sintel}, FlyingChairs \cite{dosovitskiy2015flownet} and FlyingThings3D \cite{mayer2016large} for pre-training. Such datasets use synthetic chairs and lack photorealism and realistic 3D motion. Recent advances in dataset generation -- AutoFlow \cite{AutoFlow} learns to render hyperparameters for generating synthetic flow datasets reducing the End Point Error (EPE). AutoFlow lacks 3D motion and photorealism as it utilizes a simple 2D layered model. Having  photo-realistic data, Kubric \cite{kubric} tends to outperform AutoFlow, but only on Sintel \textit{Clean} samples. %\st{Kubric \cite{kubric}, a photorealistic blender based scalable dataset generator, further reduce EPE but only on Sintel \textit{Clean} dataset but not the \textit{Final} pass.} 
This is the rendered output that contains shading but no image degradation whereas the \textit{Final} pass includes motion blur, defocus blur and atmospheric effects. Since Kubric lacks both volumetric features (like fog, mist, and rain) and advanced camera features (such as depth of field and motion blur), AutoFlow performs better even without photorealism as images are generated with motion blur and fog models as well as data augmentation for visual effects.

We compare and analyze optical flow predictions using RAFT on Sintel \textit{Clean} and \textit{Final} pass which are trained on different datasets. Table \ref{tab:flow-comparison} shows the EPE in optical flow. Note that WorldGen causes a significant increase in optical flow accuracy as compared to FlyingChairs. Furthermore, WorldGen outperforms AutoFlow due to the difference in photorealism. Note that WorldGen slightly improves over Kubric. We speculate one of the reasons to be WorldGen rendering multiple instances of the same object configuration by warping different textures on the objects, ensuring the network avoids over-fitting to object textures, and thereby learning the geometrical/motion properties of the scene. Since WorldGen supports dynamic lighting, depth of field, and motion blur modeling,  we see a significant improvement over Kubric in the \textit{Final} pass. Note that AutoFlow performs better in \textit{Final} pass (although only slightly) because the hyperparameters of AutoFlow have been learned to optimize the performance on the Sintel dataset which gives AutoFlow an unfair advantage.

\begin{table}[]
    \centering
        \caption{Optical Flow EPE Comparison of Training RAFT \cite{teed2020raft} On Different Datasets. Lower Is better.}
    \begin{tabular}{c c c c c}
        \toprule
        \multirow{2}{*}{Dataset} & \multirow{2}{*}{Dimensionality} & \multirow{2}{*}{Parameters} & Sintel & Sintel \\
          &  &  & Clean & Final  \\
         \midrule
         FlyingChairs \cite{dosovitskiy2015flownet} & 2D & Manual & 2.27 & 3.76 \\
         Kubric \cite{kubric} & 3D & Manual & 1.89 & 3.02 \\
         AutoFlow \cite{AutoFlow} & 2D & Learned & 2.08 & \textbf{2.75} \\
         WorldGen (Ours) & 3D & Manual & \textbf{1.86} & 2.87 \\
    \bottomrule
    \end{tabular}
    \label{tab:flow-comparison}
\end{table}

\subsection{Computational Photography}
Computational photography is one of the key areas that require a large collection of image data for the prediction of depth maps, denoising image data, etc. Most of the current photo-realistic synthetic data generators in robotics and computer vision do not explore and support many features that are offered by Blender, Unity, or Unreal Engines such as volumetric effects and variability in sensor sizes, camera lens, and rolling shutter sensors. Since WorldGen supports generating data for different light conditions, different \textit{bokeh} blur due to depth of field effect, motion blur, variable focal lens and sensor size along with \textit{Albedo}, \textit{Clean} and \textit{Final} renders, it can be used to generate data for the prediction of various computational photography applications such as HDR+ datasets \cite{gharbi2017deep}, depth from defocus \cite{carvalho2018deep, gur2019single}, recovering image from a rolling shutter blur \cite{cao2022learning, wang2022neural}, extracting a video sequence from a single blurred frame \cite{jin2018learning} and for learning synthesis motion blur from a pair of images \cite{brooks2019learning}. Furthermore, this work can be utilized to generate datasets for learning depth estimators using a coded aperture \cite{levin2007image,bacca2021deep} camera. We intend to add a toolbox to generate coded \textit{blurred} frames in the WorldGen environment after release.

% - learning in low light (HDR+) with a burst of photos with moving subject with its ground truth image of a high quality final shot
% \subsubsection{Estimating Depth from Motion Blur}
% - Albedo, Clean and Final Pass
% - Remove blur from the rolling shutter effect images (variable image capture time and sensor size)
% \subsubsection{Estimating Depth from Defocus}

\subsection{View Synthesis using Neural Radiance Fields}
Neural volume rendering has exploded in the last two years \cite{mildenhall2021nerf}. One of the major challenges in \cite{mildenhall2021nerf} is to synthesize in the case of dynamic relighting. These methods underperform with views ranging from satellite level capturing the entire city to the ground level, majorly due to large camera displacement and dynamic lighting. Recent advances in multi-scale scene rendering \cite{xiangli2022bungeenerf} and dynamic irradiance view synthesis \cite{mari2022sat} have adapted to work in such adverse conditions. In \cite{mari2022sat}, the authors utilize multi-date images due to their significant changes in appearance, mainly due to varying shadows and transient objects like cars and vegetation using the WorldView-3 dataset. WorldGen can generate thousands of 3D satellite views with dynamic lighting at a higher resolution than the satellite images with known camera intrinsic and extrinsic for more robust view synthesis in satellite images with larger displacement in different lighting and weather conditions (See Fig. \ref{fig:Nerf}). Also, this can be extended towards learning high-level planners for aerial cinematography \cite{mademlis2019high}.

\subsection{Active and Interactive Perception}
\label{Active-Interactive}
Not only is WorldGen an automatic 3D scene and dataset generation tool but it also serves as a framework for active and interactive perception applications. Robots such as quadrotors are often required to use the current prediction and annotations as input to the next state for its solution and testing. Currently, there is no test bench to verify the robotics algorithms that rely on active and interactive perception approaches \cite{GapFlyt, nudgeseg}. WorldGen will serve as a benchmark pseudo-simulator to validate the performance of such algorithms for better reproducibility. We hope WorldGen will become the OpenAI Gym \cite{brockman2016openai} for Active and Interactive perception.

\subsection{Generating Real World Traffic}
The problem of generating \textit{realistic} traffic scenes autonomously is a challenging task. Existing methods typically are driven by planning and reinforcement algorithms that require current and previous traffic states, traffic signal conditions, and road/pathways vertices\cite{yuan2021survey}. Since WorldGen has the ability to export these parameters, neural autoregressive models such as  \cite{tan2021scenegen} can be deployed for realistic traffic distribution. The recent advances in neural fields such as Panoptic Neural Fields \cite{KunduCVPR2022PNF} have pushed the boundary to detect traffic in a 3D representation, allowing us to perform 3D scene editing. Using such methods, WorldGen can also import real-life traffic into its model for generating photorealistic real-world city traffic scenarios.

\subsection{Human Pose Estimation}
In recent years, 3D human pose estimation has made great improvements which have been driven by large-scale datasets. However, these datasets lack accuracy in the pose as it mostly annotated by humans from open-world datasets \cite{andriluka14cvpr} that suffer from human labeling accuracy, mainly due to occlusions. On the contrary, motion capture systems have helped to generate accurate human pose data but they lack an open world environment due to motion capture space restrictions \cite{NIPS2016_35464c84}. Lastly, the simulation environment for dataset generation lacks photorealism \cite{ebadi2021peoplesanspeople}. We plan to develop another generative environment with WorldGen by utilizing Mixamo \cite{blackman2014rigging}, an open platform with a variety of character rig models and animations; and warping different 3D materials to generate a large of photorealistic character animation even in the backgrounds such as small towns and city. This will push the boundaries for monocular human pose estimation and open new avenues for special sensors such as event cameras.

% \textbf{- human \cite{} estimation in event camera (fast moving objects..Vicon is not good enough. can create human dancing in these cities etc.)}

\section{Conclusion}
\label{sec:conc}
We introduce WorldGen, a modular, generative, open-source Python API as a large-scale generative simulator with a myriad of scenes. We can generate accurate ground truth labels for optical flow, depth, surface normals, depth cameras, semantic maps and event data for driving scenes, moving object piles and object fragmentation. WorldGen also simulates camera properties, motion properties for a photorealistic perception-centric simulation and data generation. Not only can WorldGen change objects, lighting and scenes dynamically, but it can also edit objects dynamically aiding in the generation of data to better enable the generalization of neural networks such as optical flow as shown in our experiments. Currently, sensors such as SONAR and LIDAR are missing from WorldGen along with behavior modeling for self-driving cars, rigging of human characters is not implemented and we see this as a potential direction for future work.

\clearpage
\bibliographystyle{unsrt}
\bibliography{Ref}

\end{document}